%% file: neurips_2025.tex
\documentclass{article}

 \usepackage[dblblindworkshop, final]{neurips_2025}

\usepackage[hyphens]{url}
\usepackage{hyperref}
\usepackage[utf8]{inputenc} 
\usepackage[T1]{fontenc}    
\usepackage{url}            
\usepackage{booktabs}       
\usepackage{amsfonts}       
\usepackage{amsmath}
\usepackage{amssymb}
\usepackage{amsbsy}
\usepackage{mathrsfs}
\usepackage{nicefrac}       
\usepackage{microtype}      
\usepackage{xcolor}         
\usepackage{graphicx}
\usepackage{xcolor}

\usepackage{subfigure}
\usepackage{subcaption}

\newcommand{\tensor}[1]{\boldsymbol{\mathscr{#1}}}

\title{Surrogate Modeling for the Design of Optimal Lattice Structures using Tensor Completion}

\workshoptitle{AI4Mat: AI for Accelerated Materials Design}

\author{
  \begin{minipage}[t]{0.45\textwidth}
    \centering
    Shaan Pakala\\
    \normalfont University of California, Riverside\\
    \texttt{spaka002@ucr.edu}
  \end{minipage}
  \And
  \begin{minipage}[t]{0.45\textwidth}
    \centering
    Aldair E. Gongora\\
    \normalfont Lawrence Livermore National Laboratory\\
    \texttt{gongora1@llnl.gov}
  \end{minipage}
  \AND
  \begin{minipage}[t]{0.45\textwidth}
    \centering
    Brian Giera\\
    \normalfont Lawrence Livermore National Laboratory\\
    \texttt{giera1@llnl.gov}
  \end{minipage}
  \And
  \begin{minipage}[t]{0.45\textwidth}
    \centering
    Evangelos E. Papalexakis\\
    \normalfont University of California, Riverside\\
    \texttt{epapalex@cs.ucr.edu}
  \end{minipage}
}

\begin{document}

\maketitle

\input{000abstract}
\input{010introduction}
\input{020methods}
\input{030experiments}
\input{040conclusion}

\begin{ack}
\small{
S.P. \& E.E.P. were supported by National Science Foundation CAREER grant no. IIS 2046086  and CREST Center for Multidisciplinary Research Excellence in CyberPhysical Infrastructure Systems (MECIS) grant no. 2112650. A.E.G. \& B.G. contributed to this work under the auspices of the U.S. Department of Energy by Lawrence Livermore National Laboratory under Contract DE-AC52-07NA27344. IM: LLNL-JRNL-2022450.
}
\end{ack}

\bibliographystyle{plain}
\bibliography{refs}

\input{050appendix}


\end{document}

%% file: 000abstract.tex
\begin{abstract}

When designing new materials, it is often necessary to design a material with specific desired properties. Unfortunately, as new design variables are added, the search space grows exponentially, which makes synthesizing and validating the properties of each material very impractical and time-consuming. In this work, we focus on the design of optimal lattice structures with regard to mechanical performance. Computational approaches, including the use of machine learning (ML) methods, have shown improved success in accelerating materials design. However, these ML methods are still lacking in scenarios when training data (i.e. experimentally validated materials) come from a non-uniformly random sampling across the design space. For example, an experimentalist might synthesize and validate certain materials more frequently because of convenience. For this reason, we suggest the use of tensor completion as a surrogate model to accelerate the design of materials in these atypical supervised learning scenarios. In our experiments, we show that tensor completion is superior to classic ML methods such as Gaussian Process and XGBoost with biased sampling of the search space, with around 5\% increased $R^2$. Furthermore, tensor completion still gives comparable performance with a uniformly random sampling of the entire search space.

\end{abstract}

%% file: 010introduction.tex
\section{Introduction}

Efficiently designing new materials with optimal properties can be a significant challenge due to the explosive number of combinations as new design variables are introduced. When searching for new materials with particular material property values, it can be very expensive to exhaustively generate these combinations of materials. For this reason, there is growing interest in being able to predict these material property values, without having to first produce the material. Computational methods have shown great promise for inferring material property values, allowing material scientists to quickly find materials with optimal properties. This includes the use of Density Functional Theory (DFT) \cite{hafner2006toward,makkar2021review,schleder2019dft,verma2020status}, machine learning (ML) \cite{alberi20182019,gongora2024accelerating,haghshenas2024full,hu2024realistic,li2024md,pakala2025tensor,wang2020accelerated,zhuo2018predicting}, or both \cite{seko2014machine}. In this work, we focus on designing optimal lattice structures, with regards to mechanical performance \cite{gongora2024accelerating, gongora2021designing}.

Unfortunately, these ML methods still underachieve in scenarios where the training data does not come from a uniformly random sample of the entire search space. This scenario may arise in practice when an experimentalist might synthesize and validate certain materials of the design space more than others, due to convenience. In order to overcome this, we abstract away the various components of material design and instead model them as instances of tensor completion. Here, each design parameter to be optimized corresponds to an individual mode of our tensor, so we can leverage tensor completion algorithms to predict the entirety of the search space. In our experiments, tensor completion methods provide nearly state-of-the-art performance in typical supervised learning scenarios. Furthermore, we simulate an experimentalist performing experiments to serve as training data by unevenly sampling across design variables. In these scenarios, tensor completion gives better results than other baseline ML methods, including Gaussian Process \cite{williams1995gaussian}, XGBoost \cite{chen2016xgboost}, and some ensembles of multiple methods. Of these baseline methods, Gaussian Process performed the best in our experiments so we show results compared against this. Additionally, we make our code public at \url{https://github.com/shaanpakala/Surrogate-Modeling-for-the-Design-of-Optimal-Lattice-Structures-using-Tensor-Completion}.

%% file: 020methods.tex
\section{Methods}

In this section we give a detailed description on the methods used in this work, including our pipeline for surrogate modeling with tensor completion, its training, and the dataset used for evaluation.

\subsection{Tensor Completion for Surrogate Modeling}

\begin{figure}[h]
    \centering
    \begin{center}
        \includegraphics[width = 0.7\textwidth]{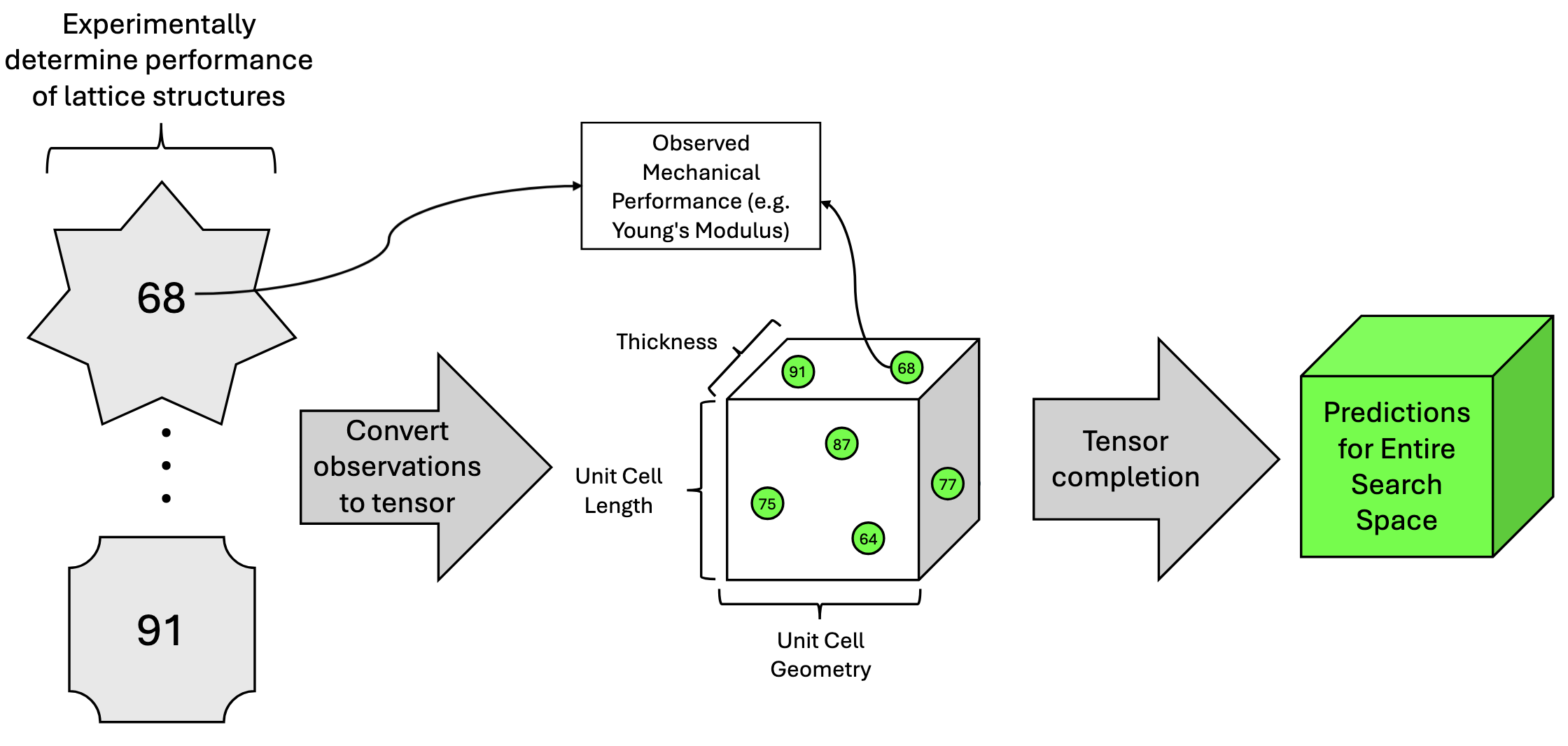}
        \caption{We present various material design problems as instances of tensor completion, in order to abstract away the various design components and use tensor methods to infer the entire search space. \label{overview_figure}}
    \end{center}
\end{figure}

In Figure \ref{overview_figure}, we show the complete surrogate modeling pipeline we are proposing. After experimentally determining the performance for various material structures, we can convert those observations to entries of a tensor and use tensor completion to infer the performance of the remainder of the design search space.

\subsubsection{Tensors \& Tensor Decomposition}

Tensors are a general term for multidimensional arrays. In other words, a 1st order tensor is just a vector, and a 2nd order tensor is a matrix. In this work, we will be looking at higher order tensors, denoted $\tensor{X}$. Tensor decomposition is a general term for expressing a tensor as several smaller factors, and is an extension of matrix factorization to multidimensional datasets. A common form of tensor decomposition is the Canonical Polyadic Decomposition (CPD) 
\cite{kolda2009tensor,sidiropoulos2016tensor}. CPD expresses a tensor as a sum of rank-one tensors. A rank $R$ CPD decomposition of a third-order tensor $\tensor{X} \in \mathbb{R}^{I \times J \times K} $ would be expressed as:
$
\tensor{X} \approx \sum_{r=1}^{R} (\mathbf{a}_r \circ \mathbf{b}_r \circ \mathbf{c}_r),
$
where $\circ$ denotes outer product, $\mathbf{a}_r \in \mathbb{R}^I, \mathbf{b}_r \in \mathbb{R}^J, \text{ and } \mathbf{c}_r \in \mathbb{R}^K$.

Tensor decomposition is a powerful tool, not only for data compression but also for analyzing multidimensional datasets. For example, the smaller tensor factors are commonly used for pattern discovery in these multidimensional datasets, and further more for tensor completion. Tensor completion is the process of filling in the missing values of a tensor.

\subsubsection{Tensor Completion Training}

\begin{figure}[h]
    \centering
    \begin{center}
        \includegraphics[width = 0.6\textwidth]{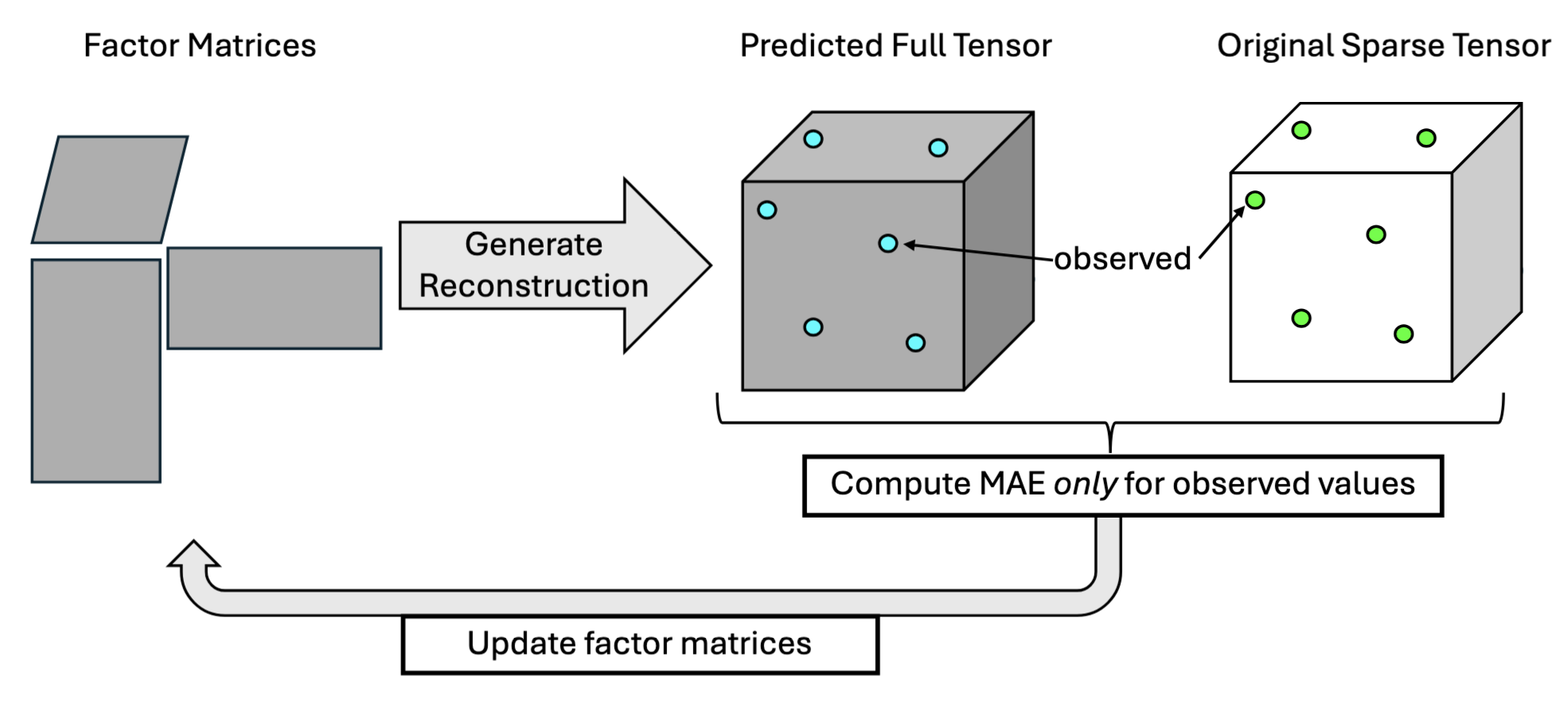}
        \caption{Here we visualize the training process, where we iteratively update our factor matrices to produce a better tensor reconstruction. The gray regions correspond to unobserved values. \label{tensor_completion_training}}
    \end{center}
\end{figure}

We illustrate the training of our tensor completion models in Figure \ref{tensor_completion_training}.  We first randomly initialize the factor matrices to generate an initial (random) reconstruction of the full tensor. Then we use error of the observed tensor values and the associated predictions to iteratively get a more accurate reconstruction of the full tensor. For our error function, we use Mean Absolute Error (MAE) = $\frac{1}{n}\sum_{i=1}^{n}|y_i - \hat y_i|$. To optimize the parameters (factor matrices) with regards to this error term, we use the Adam optimizer \cite{kingma2014adam}.

\subsubsection{Tensor Completion Models}

There are several classes of tensor completion methods. There are classical methods such as CPD \cite{kolda2009tensor} and TuckER \cite{balazevic2019tucker}, and there are more advanced neural tensor completion methods such as NeAT \cite{ahnneural} or CoSTCo \cite{liu2019costco}, which leverage the use of neural networks in addition to tensor decomposition.

For our experiments, we use CPD-S (a CPD-based method that imposes smoothness constraints on some of the tensor factors) \cite{ahn2021time, 10825934}, NeAT \cite{ahnneural}, and an ensemble tensor completion method to aggregate the results of multiple tensor completion algorithms. For our results, we only display an ensemble of CPD-S with different ranks, and a NeAT model as it showed the best performance on this dataset.

\subsection{Baseline Comparisons}

In this work, we compare our tensor completion methods against Gaussian Process, XGBoost, Multi-Layer Perceptrons (MLPs), and using an ensemble of ML methods. Due to limited space, we only show the results for Gaussian Process, as it showed the best results of the baselines on this dataset.

\subsection{Dataset Description}
 
To verify the utility of using tensor completion as a surrogate model for lattice structure design, we use the dataset from Gongora et al. \cite{gongora2024accelerating} with various lattice structures' design components and the corresponding mechanical performance. Mechanical performance is characterized using the Young's modulus ($E$) values and the ratio between $E$ and the design's mass $m$ ($\tilde{E}$). These are essential mechanical properties to describe the elastic behavior of a structure under an applied force \cite{gongora2024accelerating}. 

%% file: 030experiments.tex
\section{Experiments}

In this section, we experimentally validate our method in a variety of settings, with uniformly random and biased random sampling of the trainset, with varying training sizes.

\subsection{Tensor Completion Performance}

\begin{figure}[!h]

    \centering
    \subfigure[Parity Plot for Testing Values]{
        \includegraphics[width=0.32\linewidth, height = 3.5cm]{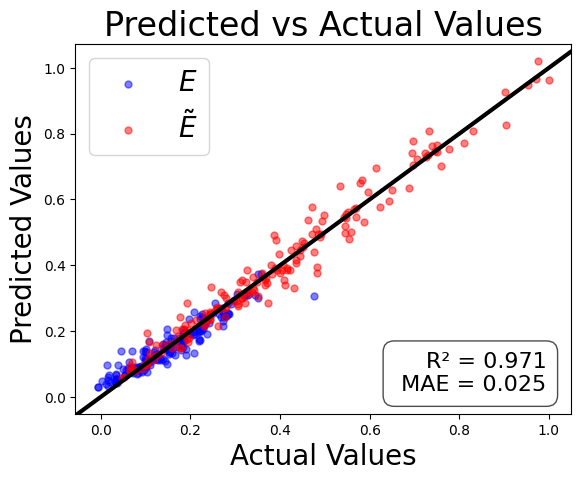}
    }
    \subfigure[Efficiency Analysis]{
        \includegraphics[width=0.42\linewidth, height = 3.5cm]{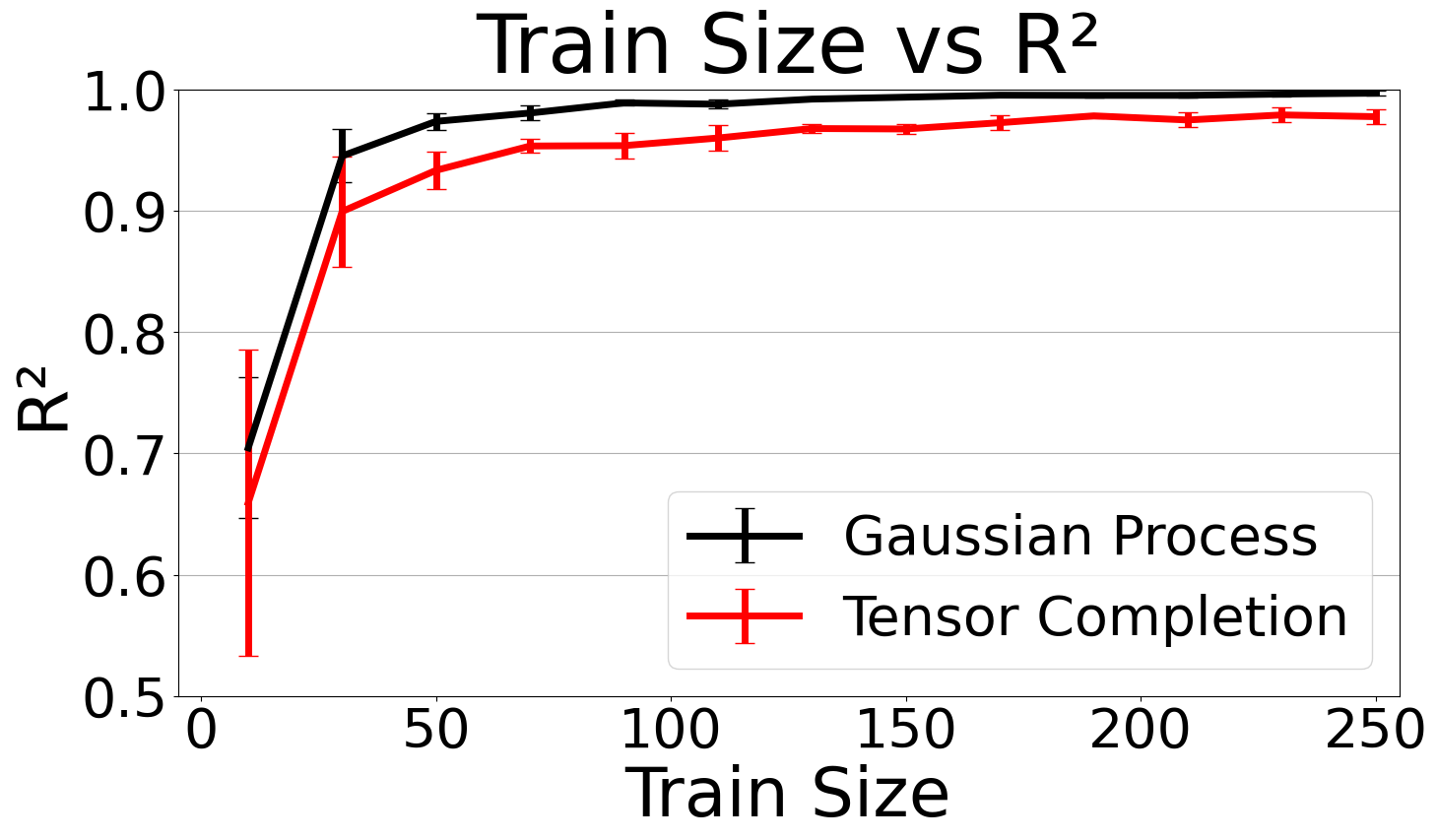}
    }

    \caption{In (a) we display a parity plot for predicting $E$ and $\tilde{E}$ simultaneously. We use 100 train and 170 test values. In (b) we display the $R^2$ for a uniform random sampling of different training sizes. We show the mean and standard deviation on the test set, using 5 iterations for each train size. \label{supervised_learning_performance}}

\end{figure}

From Figure \ref{supervised_learning_performance} we see tensor completion gives good results in typical supervised learning tasks (i.e. uniform random sampling). However, we achieve better results with less data with Gaussian Process.

\subsection{Biased Sampling}

Here we observe surrogate modeling performance when the training data is not uniformly randomly sampled throughout the search space. We sample training values unevenly across different values of a certain design variable. The goal is to simulate an experimentalist who might have conducted more experiments in a specific region of the search space, instead of uniformly randomly. This is useful if a certain region of the search space is cheaper or more convenient (e.g. with regards to time or money) to experimentally validate.

\begin{figure}[!h]

    \centering
    \subfigure[Example of biased sampling]{
        \includegraphics[width=0.35\linewidth, height = 3.25cm]{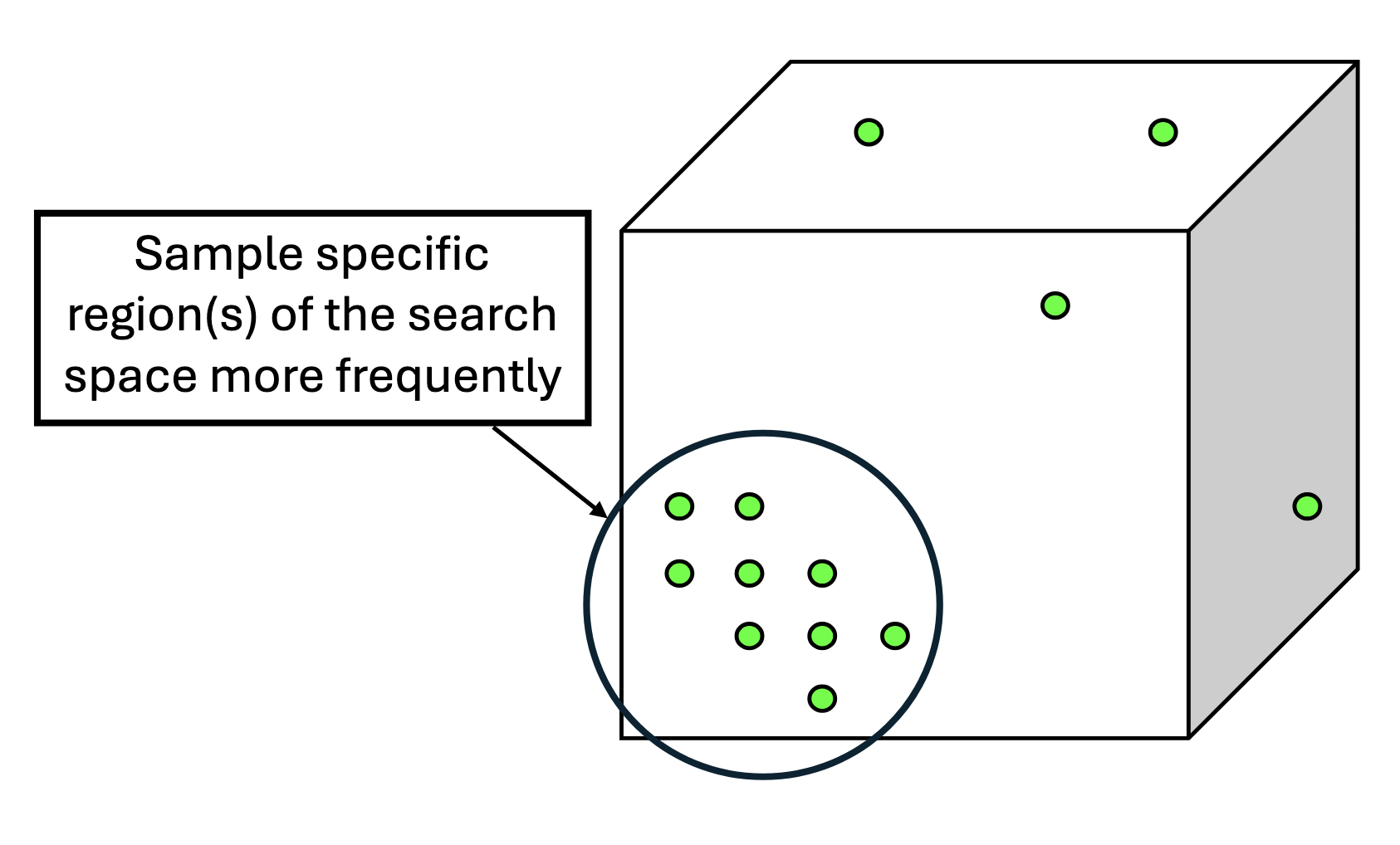}
    }
    \subfigure[R$^2$ value for biased sampling experiments]{
        \includegraphics[width=0.45\linewidth, height = 3.25cm]{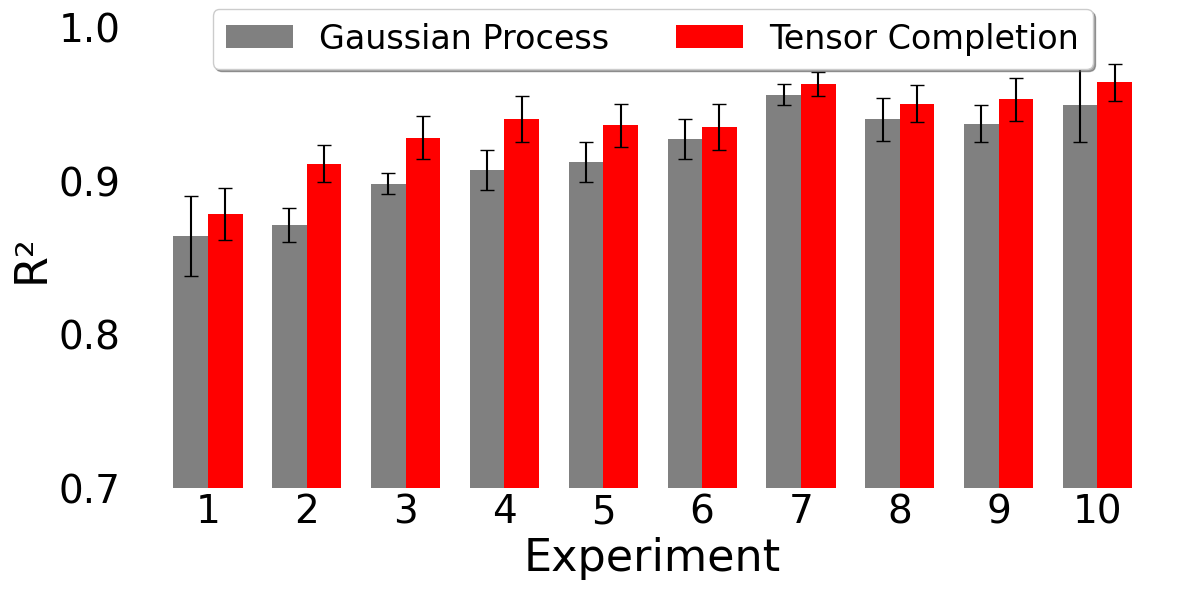}
    }

    \caption{In (a) we visualize how the search space might be sampled in a biased manner. In (b), we show results for a biased-random sampling of the search space. We display the average and standard deviation of the $R^2$ for 5 iterations. Going from experiment 1 to 10, we decrease the bias in sampling by making the range smaller. The exact setup is provided in the appendix section in Table \ref{biased_sampling_trainset}.  \label{biased_sampling_results}}

    \label{data_augmentation_aug_size_fig}

\end{figure}

From Figure \ref{biased_sampling_results} we observe that tensor completion methods are better in handling a biased sampling in the training set, and less prone to overfitting to the observed values. This makes tensor completion very useful for a wider variety of real-world scenarios than standard supervised learning techniques.




%% file: 040conclusion.tex
\section{Conclusion}

In this work, we express the design of optimal lattice structures (with respect to mechanical performance) as a tensor completion problem. This can handle both typical (uniformly random sampling) and atypical (biased sampling) supervised learning problems to accelerate the design of optimal lattice structures. For these reasons, tensor completion could be promising as a surrogate model to accelerate the search for optimal lattice structures with an exponentially increasing design space.

%% file: 050appendix.tex
\appendix

\section{Appendix}

\subsection{Limitations}

The main limitation of this work is considering typical supervised learning scenarios, when the training data comes from a uniformly random sampling of the entire data or search space. In our experiments, we see that a Gaussian Process exhibits superior performance in these scenarios.

\subsection{Experimental Details}

All code used for experiments can be found at   \url{https://github.com/shaanpakala/Surrogate-Modeling-for-the-Design-of-Optimal-Lattice-Structures-using-Tensor-Completion}. However, we cannot release the dataset due to the dataset owner's institutional regulations. More dataset details can be found on the original paper \cite{gongora2024accelerating}.

\subsubsection{Hardware}

Experiments were fairly lightweight to run. They all were conducted using a MacBook Air 2020 laptop with an Apple M1 chip (8GB).

\subsubsection{ML Models}

For our experiments, we used an ensemble tensor completion method consisting of a NeAT Rank 24 \& 32, CPD Ranks 1, 2, \& 4, and CPD-S Ranks 1, 2, \& 4. Each model was trained using Adam optimizer \cite{kingma2014adam} with a learning rate and weight decay of 0.01, using MAE as the objective function. A random forest regression using 100 trees was used to aggregate the results of the tensor models.

For our baseline methods, we tried a variety of XGBoost, MLPs, and Gaussian Process models using Scikit-Learn \cite{scikit-learn}. The best of the baselines (and only one shown in the experiments due to space constraints) was a Gaussian Process with kernel = ConstantKernel(1.0, (1e-3, 1e3)) $\times$ RBF(1.0, (1e-2, 1e2)) + WhiteKernel(1e-3, (1e-5, 1e1)) using Scikit-Learn's Gaussian Process kernel functions, and an alpha value = 0.01. These were determined empirically by comparing the performance of various hyperparameter values.

\subsubsection{Section 3.2 Biased Sampling Details}
\begin{table}[h!]
\centering
\small
\begin{tabular}{|l|cccccc|}
\hline
Experiment & Gyroid & Schwarz & Diamond & Lidinoid & Split P & Range \\
\hline
1 & 40 & 21 & 7 & 6 & 3 & [3, 40] \\
2 & 14 & 21 & 5 & 5 & 40 & [5, 40] \\
3 & 8 & 23 & 7 & 15 & 40 & [7, 40] \\
4 & 9 & 24 & 12 & 19 & 40 & [9, 40] \\
5 & 11 & 21 & 16 & 40 & 25 & [11, 40] \\
6 & 14 & 13 & 40 & 22 & 35 & [13, 40] \\
7 & 36 & 15 & 28 & 40 & 16 & [15, 40] \\
8 & 40 & 17 & 21 & 28 & 37 & [17, 40] \\
9 & 30 & 35 & 29 & 19 & 40 & [19, 40] \\
10 & 34 & 40 & 32 & 39 & 21 & [21, 40] \\
\hline
\end{tabular}

\caption{Number of training samples used for each experiment in Figure \ref{biased_sampling_results}. Each unit cell geometry type (i.e. tensor slice) has 54 total values (54 values $\times$ 5 slices = 270 total values in search space). For each experiment, the number of training samples for each unit cell geometry type comes from an exponential distribution with scale = 1. This was to create a high disparity in the number of samples for each value. We then converted this exponential distribution to be in the range [$l$, 40], where $l = 3 + 2 \times e_{num}$ ($e_{num}$ is the current experiment number), and then rounded to the nearest integer value. This was to iteratively decrease the bias in sampling, by making the range of values smaller (i.e. increasing the lower bound $l$) as we go from experiment 1 to experiment 10. We start with experiment 1 from an exponential distribution on the range [3, 40], and we end with experiment 10 on the range [21, 40]. \label{biased_sampling_trainset}}
\end{table}